\documentclass[runningheads]{llncs}
\usepackage[T1]{fontenc}
\usepackage{graphicx,verbatim}
\usepackage{amsmath}
\usepackage{bm}
\usepackage{hyperref}
\usepackage{color}

\begin{document}
\title{BioFact-MoE: Biologically Factorized Mixture of Experts for Vision–Language Prognostic Modeling in Hepatocellular Carcinoma}
\titlerunning{BioFact-MoE for Vision–Language Prognostic Modeling}
%
\author{Junlin Yang\inst{1} \and
Tian Yu\inst{2} \and
Nicha C. Dvornek\inst{1,2} \and Yuexi Du\inst{2} \and Peiyu Duan\inst{2} \and Annabella Shewarega \inst{1} \and Lawrence H. Staib \inst{1,2,3} \and James S. Duncan \inst{1,2,3,4} \and Julius Chapiro \inst{1} }
\authorrunning{F. Author et al.}
%
\institute{Department of Radiology \& Biomedical Imaging, \and
Department of Biomedical Engineering, 
\and 
Department of Electrical Engineering, 
\and 
Department of Statistics \& Data Science
\\ Yale University, New Haven, CT, 06510, USA
\\
\email{junlin.yang@yale.edu}}


  
\maketitle              
%


  
%
\begin{abstract}

Hepatocellular carcinoma (HCC) is biologically heterogeneous, shaped by the interplay between hepatic functional reserve and tumor-related oncologic factors; thus, similar survival outcomes may reflect fundamentally different underlying biological processes. Prognostic modeling in HCC is informed by rich multimodal information from multiparametric MRI and radiology reports from routine clinical practice. Existing prognostic vision–language models (VLMs) learn a single entangled latent representation that blends hepatic and tumor-related factors, limiting both accuracy and biological interpretability. We present BioFact-MoE, a biologically factorized Mixture of Experts (MoE) framework that explicitly decomposes liver and tumor factors via biologically supervised experts within a residual MoE survival architecture. On a HCC cohort of N=588 patients (pretrained on 4,582 3D MRI image-report pairs), BioFact-MoE consistently improves survival prediction over all baselines across time horizons, achieving 12-, 18-, and 24-month AUCs of 75.33\%, 75.85\%, and 73.96\%. Beyond scalar risk prediction, gated expert weights enable phenotype-aware risk stratification. Pathway-informed gating uncovers clinically meaningful treatment-associated survival heterogeneity. In held-out validation, hepatic and tumor embeddings show selective associations with liver function and tumor burden markers, respectively (p<0.05), without supervision. The code is available at https://github.com/jy-639/BioFact-MoE.



\end{abstract}
\section{Introduction}

Hepatocellular carcinoma (HCC) is a leading cause of cancer-related mortality~\cite{foglia2023hepatocellular}, with prognosis governed by two
interacting yet biologically distinct axes: hepatic functional reserve and
tumor-specific oncologic behavior~\cite{liu2016prognosis}.
Patients with similar survival can have different
underlying drivers. In some patients, poor outcomes are driven by severe hepatic dysfunction, whereas in others they are attributable to tumor biology. 
Disentangling these pathways is critical for risk attribution and
personalized treatment planning, yet most existing data-driven prognostic models compress both
axes into a single entangled representation. Recent vision-language models (VLMs) improve radiology representation learning through image–report alignment, and finer-grained semantic grounding
at the sentence, entity, and concept level has been explored~\cite{du2025multi,zhang2026medground,li2025more,wu2025concept}.
However, these approaches organize representations by semantic granularity rather than clinically grounded biological pathways, implicitly mixing hepatic and tumor signals, limiting prognostic modeling performance and biological interpretability in heterogeneous cohorts such as HCC. In clinical practice, radiology reports routinely describe hepatic functional reserve and tumor-related characteristics. Yet standard representation learning does not explicitly disentangle these factors. We therefore introduce pathway-specific structure as a domain-informed inductive bias to encourage biologically aligned representations.
Parameter-efficient adaptation via LoRA~\cite{hu2022lora} provides a mechanism for lightweight specialization, but a single shared adapter applies the same transformation regardless of biological heterogeneity.
Recent LoRA–MoE approaches, including MoLoRA~\cite{zadouri2023pushing}, LoRAMoE~\cite{dou2024loramoe}, and MixLoRA~\cite{li2024mixlora}, learn expert routing and specialization implicitly from downstream task objectives, without incorporating prior structural differentiation among experts. Such designs are effective for multi-task adaptation but are less suited to settings where meaningful domain structure exists. In HCC prognostic modeling, biological axes are present in multimodal pretraining data yet are not explicitly encoded in survival labels. 

We present \textbf{BioFact-MoE} to address this gap by enforcing expert specialization through biologically defined pathway inputs during unsupervised contrastive pretraining, establishing domain-informed inductive bias before survival supervision is introduced.
\textbf{Contributions:} \textbf{(1) Biologically factorized pretraining.}
Large language model (LLM)-guided report decomposition paired with anatomical patch masking trains pathway-specific LoRA adapters that independently encode hepatic and tumor representations.
\textbf{(2) Improved prognostic performance.}
  BioFact-MoE achieves 12-, 18-, and 24-month AUCs of 75.33\%, 75.85\%, and
  73.96\%, outperforming image-only, multimodal, vision–language pretraining, and MoE baselines. 
\textbf{(3) Residual MoE survival architecture.}
  A gated residual MoE head dynamically routes patient-specific risk
  contributions across hepatic and tumor pathways, enabling phenotype-aware 
  risk stratification.
  Exploratory analysis shows the liver-dominant low-risk subgroup benefits from transarterial chemoembolization (TACE), consistent with clinical expectations.
\textbf{(4) Biological validation.}
Hepatic embeddings align with Platelet-Albumin-Bilirubin (PALBI), while tumor embeddings align with bilobar disease and immunoscore (all p<0.05), without supervision, confirming biological specificity.

\section{Method}

\begin{figure}[t]
\includegraphics[width=\textwidth]{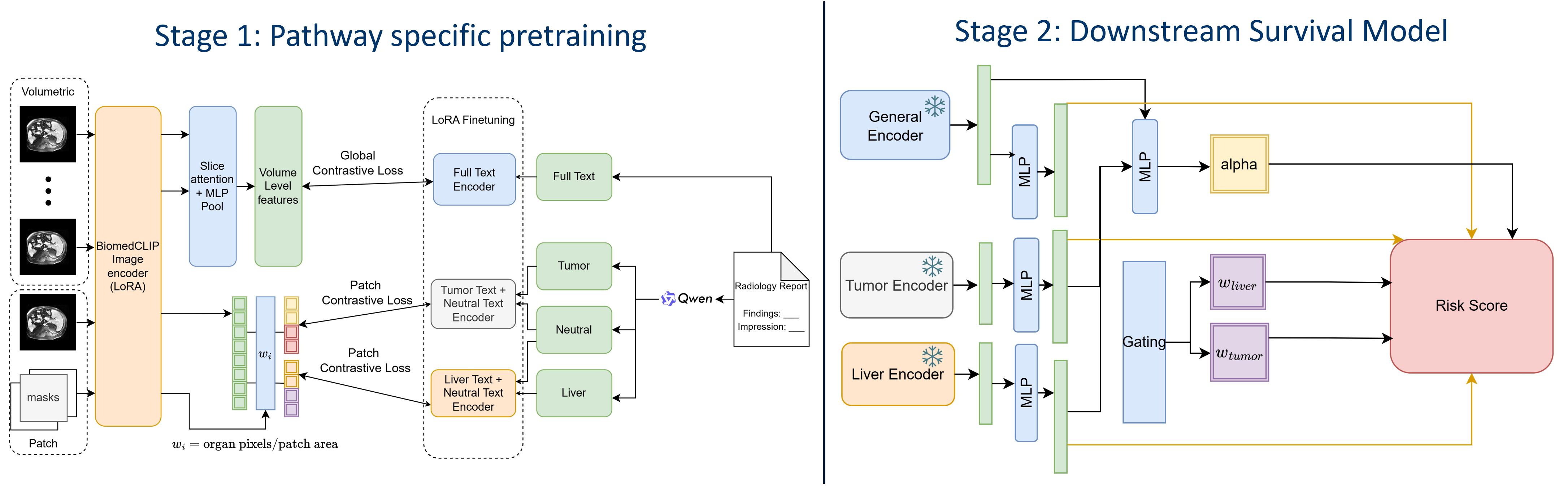}
\caption{BioFact-MoE framework.
\textbf{Stage 1:} LLM-guided report decomposition supervises three
pathway-specific LoRA adapters via contrastive pretraining with anatomical
patch masking.
\textbf{Stage 2:} Frozen pathway encoders are integrated by a residual MoE
survival head with adaptive gating for Cox-based survival prediction.}
\label{figure1,pipeline}
\end{figure}

\subsection{Stage 1: Pathway-Specific Biological Factorized Pretraining}
BioFact-MoE framework operates in two stages (Fig. \ref{figure1,pipeline}). In the first stage,  biologically factorized pathway-specific adapters are pretrained using a contrastive vision–language objective on MRI–report pairs. We build on BiomedCLIP~\cite{zhang2024biomedclip} (ViT-B/16 image encoder,
PubMedBERT text encoder), with all backbone weights frozen. For volumetric MRI, we adopt a 2.5D strategy: each axial slice is encoded
independently, a lightweight 1-layer self-attention transformer with learnable
positional embeddings aggregate inter-slice context, and a two-layer MLP
attention module produces a weighted sum of slice embeddings as the
volume-level representation.

\textbf{LLM-Guided Report Decomposition}
Each radiology report $t$ is parsed offline by Qwen~\cite{yang2025qwen3} into three segments:
$t \to (s_{\text{liver}},\, s_{\text{tumor}},\, s_{\text{neutral}})$,
where $s_{\text{liver}}$ captures hepatic parenchymal quality, portal
hypertension signs, and splenic findings; $s_{\text{tumor}}$ captures tumor
morphology, vascularity, and aggressiveness; and $s_{\text{neutral}}$ retains remaining context.
During pretraining, the neutral segment is concatenated to both
$s_{\text{liver}}$ and $s_{\text{tumor}}$, forming pathway-specific textual
inputs that preserve shared clinical context while maintaining biological
specialization.

\textbf{Anatomical Patch Masking}
Organ masks from TotalSegmentator~\cite{akinci2025totalsegmentator}
 weight patch tokens by organ occupancy
$w_i = \text{(organ pixels in patch)} \,/\, \text{(patch area)}$, so the
pathway embedding $v = \sum_i w_i t_i / \sum_i w_i$ is restricted to
pathway-relevant anatomy.
The liver pathway uses liver and spleen masks; the tumor pathway uses liver,
portal vein, and inferior vena cava (IVC).

\textbf{Pathway-Specific LoRA}
Three independent LoRA parameter sets $(A_p, B_p)$,
$p \in \{\text{general, liver, tumor}\}$ (rank $r{=}4$, $\alpha{=}8$) are
trained separately via symmetric InfoNCE loss \cite{radford2021learning}
$\mathcal{L} = \frac{1}{2}(\mathcal{L}_{\text{img}\to\text{txt}} +
\mathcal{L}_{\text{txt}\to\text{img}})$ at $\tau{=}0.1$, with AdamW at a learning rate of $1\times 10^{-4}$ for 250 epochs with a batch size 32.

\subsection{Stage 2: Residual MoE Survival Modeling}

All pathway encoder checkpoints (including LoRA parameters) are frozen.
The general encoder produces $z_{\text{base}} \in R^{1024}$.
Pathway checkpoints yield $z_{\text{liver}}, z_{\text{tumor}} \in
R^{1024}$.
Two expert MLPs map pathway embeddings to 256-dim representations
$e_{\text{liver}}, e_{\text{tumor}}$.
A core component of the MoE framework is gating function $G$ for dynamically routing patient-specific contributions across experts. In our case, $G: R^{3072} \to R^2$ is a learned linear projection.

The gate input $g_i=[z_{\text{base}};\; \delta z_i;\; |\delta z_i|]$ concatenates the global patient context $z_{\text{base}}$,
the signed inter-pathway discrepancy $\delta z_i=z_{\text{liver}} - z_{\text{tumor}}$, and its absolute magnitude
$|\delta z_i|$, allowing the gate to jointly reason about the patient's overall
profile and the degree of disagreement between hepatic and tumor representations.
The resulting soft weights $w_i = [w_{liver}^{i}, w_{tumor}^{i}] = \text{softmax}(G(g_i))$ (summing to 1) reflect the relative dominance of each pathway for patient $i$: a large $w_{liver}^{i}$
indicates a liver-dominant phenotype, while a large $w_{tumor}^{i}$ indicates a
tumor-driven one. These weights are later used for phenotype-aware patient stratification,
enabling clinically interpretable subgroup discovery beyond a scalar risk score.

A per-patient coefficient $\alpha_i = \sigma(\text{MLP}_\alpha([z_{\text{base}};
e_{\text{liver}}; e_{\text{tumor}}]))$ modulates expert contribution,
with the final risk score:
\begin{equation}
r_i = \underbrace{h_{\text{base}}(z_{\text{base}})}_{\text{base risk}}
+ \;\alpha_i\!\left(
  w_{i1}\,\delta r_i^{\text{liver}} + w_{i2}\,\delta r_i^{\text{tumor}}
\right).
\end{equation}

The MoE survival head is optimized using the Cox partial likelihood with AdamW (learning rate $1\times 10^{-4}$) for 50 epochs with a batch size of 16.


\section{Experiments}

\subsection{Datasets and Data Preprocessing}

We curated two cohorts from local hospitals. This retrospective study was IRB approved.
\textbf{Pretraining:} 4,582 abdominal MRI–report pairs from 1,896 patients, used for vision–language pretraining without survival supervision.
\textbf{Downstream:} 588 HCC patients with baseline multiparametric MRI-report pairs and longitudinal follow-up for overall survival, randomly split into 400/188 for training/test. Subsets have structured biomarkers: PALBI ($n{=}118$), ALBI ($n{=}275$), Immunoscore ($n{=}62$), and bilobar disease ($n{=}386$), none used for supervision.

MRIs were resampled to isotropic resolution, normalized, and cropped to 224×224×60. TotalSegmentator~\cite{akinci2025totalsegmentator,isensee2021nnu} organ masks were only used during pretraining. Reports were de-identified, retaining Findings and Impression sections.

\subsection{Baselines and Metrics}

We compare four categories of baselines: image-only survival modeling, multimodal fusion, state-of-the-art biomedical VLMs, and MoE architectures. Image-only and multimodal fusion baselines are trained solely on the downstream survival cohort, while VLMs and MoE models first leverage large-scale image–report pretraining. Performance is evaluated using time-dependent AUC at 12, 18, and 24 months based on three independent runs with different random seeds.

\noindent\textbf{Image-only survival}
A pretrained 2D ResNet processes axial MRI slices independently. Slice-level features are averaged to obtain a patient-level representation, which is optimized under the Cox proportional hazards objective \cite{katzman2018deepsurv}.

\noindent\textbf{Multimodal survival}
Image and Report Fusion follows standard late-fusion multimodal modeling \cite{willis2025exploring}.
A pretrained 2D image encoder processes individual axial MRI slices, 
which are averaged to obtain a patient-level embedding.
A pretrained text encoder extracts a global report
embedding. Both embeddings are concatenated and fed
into a Cox regression head for survival prediction~\cite{katzman2018deepsurv}.

\noindent\textbf{Vision-language fine-tuning.}
BiomedCLIP~\cite{zhang2024biomedclip} is fine-tuned end-to-end for
survival by appending a Cox regression head to the joint image--report
embedding.
GLoRIA~\cite{huang2021gloria} extends global contrastive learning with 
local alignment, contrasting image sub-regions with individual
report words;
 a Cox head is
applied to patch- and global-level features.
Finally, we implement an entity-aligned variant that 
exploits organ segmentation masks to explicitly
supervise alignment between report-extracted anatomical entities and
their corresponding segmented regions, providing a more robust
grounding signal.

\noindent\textbf{MoE}
SparseMoE implements a standard sparsely-gated MoE layer
 \cite{shazeer2017outrageously}. Given the frozen BiomedCLIP fused
embedding, a learned gating network
produces routing logits over $E{=}4$ experts. We select only the two highest-scoring experts
and combine them using the normalized gate weights as output.

\subsection{Survival Performance}

\begin{table}[t]
\centering
\caption{Time-dependent AUC (mean$\pm$std) at 12, 18, 24 months. \textbf{Bold} = best.}
\label{tab:comparison}
\setlength{\tabcolsep}{4pt}
\renewcommand{\arraystretch}{1.15}
\scalebox{0.9}{
\begin{tabular}{llccc}
\hline
\textbf{Category} & \textbf{Method} & \textbf{12m} & \textbf{18m} & \textbf{24m} \\
\hline\hline

Image-only & ResNet-Cox & 62.74 $\pm$ 1.39 & 64.28 $\pm$ 0.97 & 63.26 $\pm$ 0.96 \\

\hline

Multimodal survival 
& Image+Report Fusion & 64.69 $\pm$ 0.99 & 65.68 $\pm$ 1.06 & 64.19 $\pm$ 1.13\\
\hline

Vision--language & BiomedCLIP~\cite{zhang2024biomedclip} & 65.71 $\pm$ 5.87 & 67.97 $\pm$ 0.64 & 66.91 $\pm$ 2.31 \\
 & GLoRIA~\cite{huang2021gloria} & 73.20 $\pm$ 0.76 & 72.13 $\pm$ 0.88 & 70.00 $\pm$ 1.22 \\
  & Entity Alignment & 74.27 $\pm$ 1.86 & 72.05 $\pm$ 1.52 & 70.26 $\pm$ 2.39 \\
\hline

 MoE & SparseMoE  & 72.56 $\pm$ 2.37 & 73.31 $\pm$ 1.71& 71.58 $\pm$ 0.87\\
\hline

Ours & \textbf{BioFact-MoE}
& \textbf{75.33 $\pm$ 2.60}
& \textbf{75.85 $\pm$ 1.45}
& \textbf{73.96 $\pm$ 3.04} \\
\hline
\end{tabular}}
\end{table}

\begin{figure}[t]
    \centering
    \includegraphics[width=0.9\linewidth]{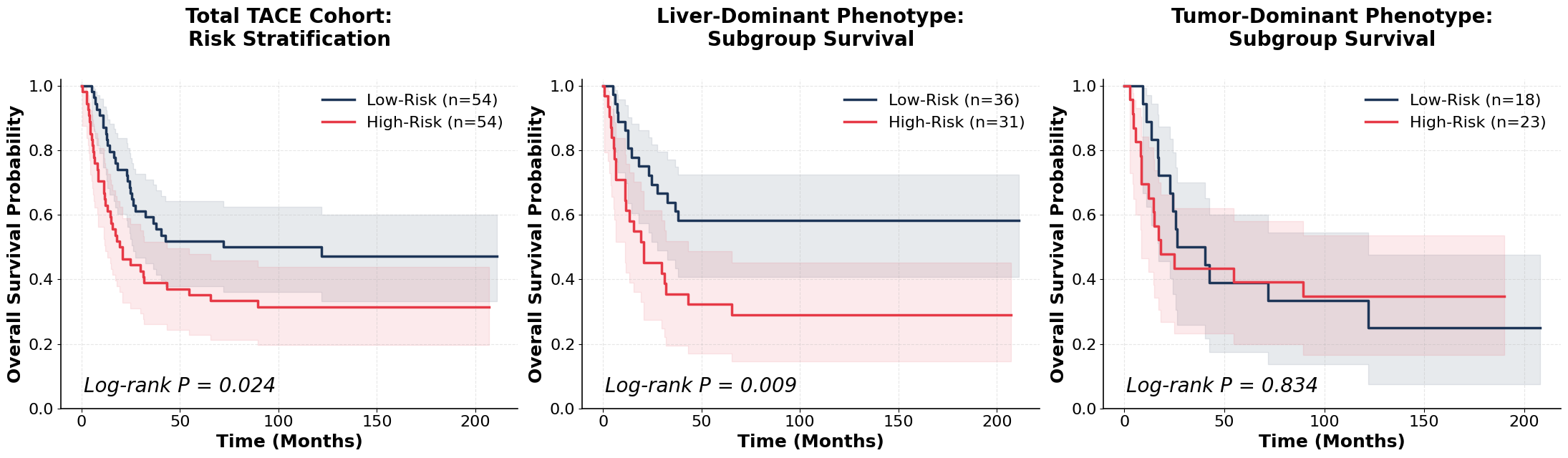}
    \caption{Phenotype-Aware Stratification: Beyond scalar risk,
gate weights stratify patients by dominant biological axis. Among patients receiving the same TACE treatment, liver-driven and tumor-driven subgroups show significantly different survival trajectories, capturing heterogeneity invisible to clinical staging alone.}
    \label{fig:phenotype-aware}
\end{figure}

\begin{figure}[t]
    \centering
    \includegraphics[width=0.8\linewidth,height=0.38\textheight,keepaspectratio]{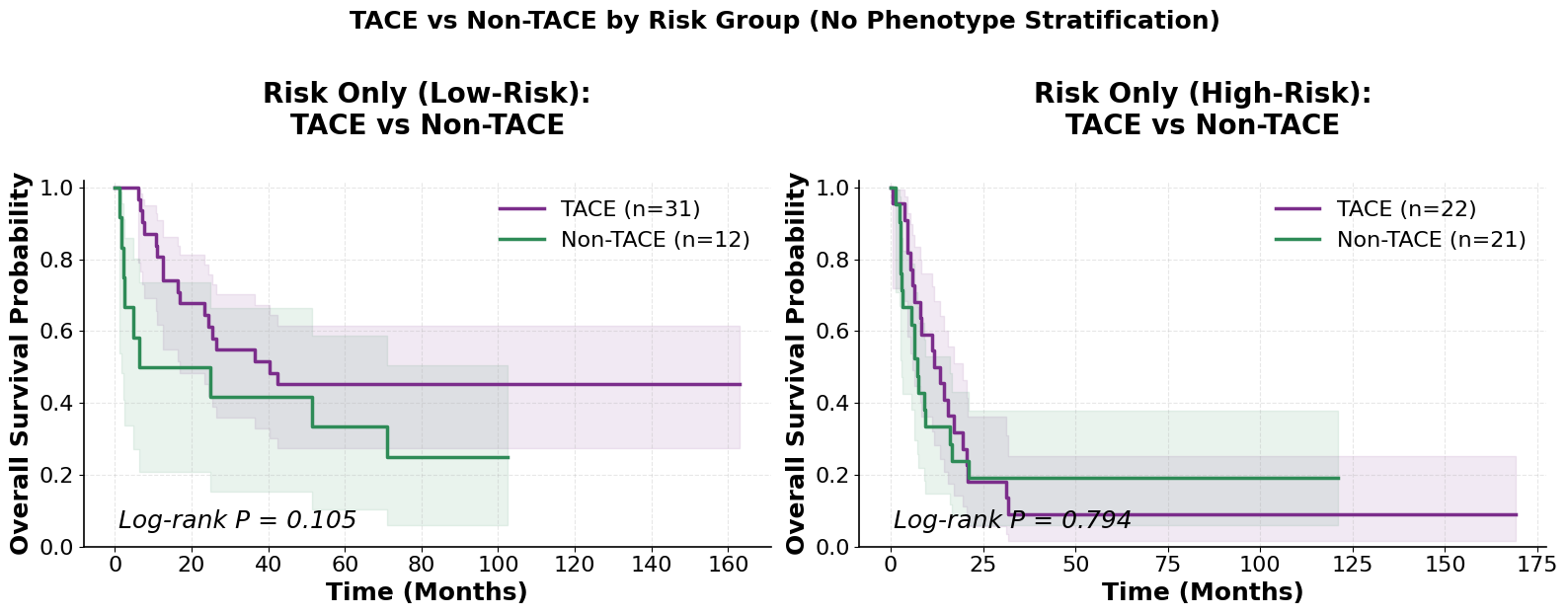}
    
    
    \includegraphics[width=0.8\linewidth,height=0.38\textheight,keepaspectratio]{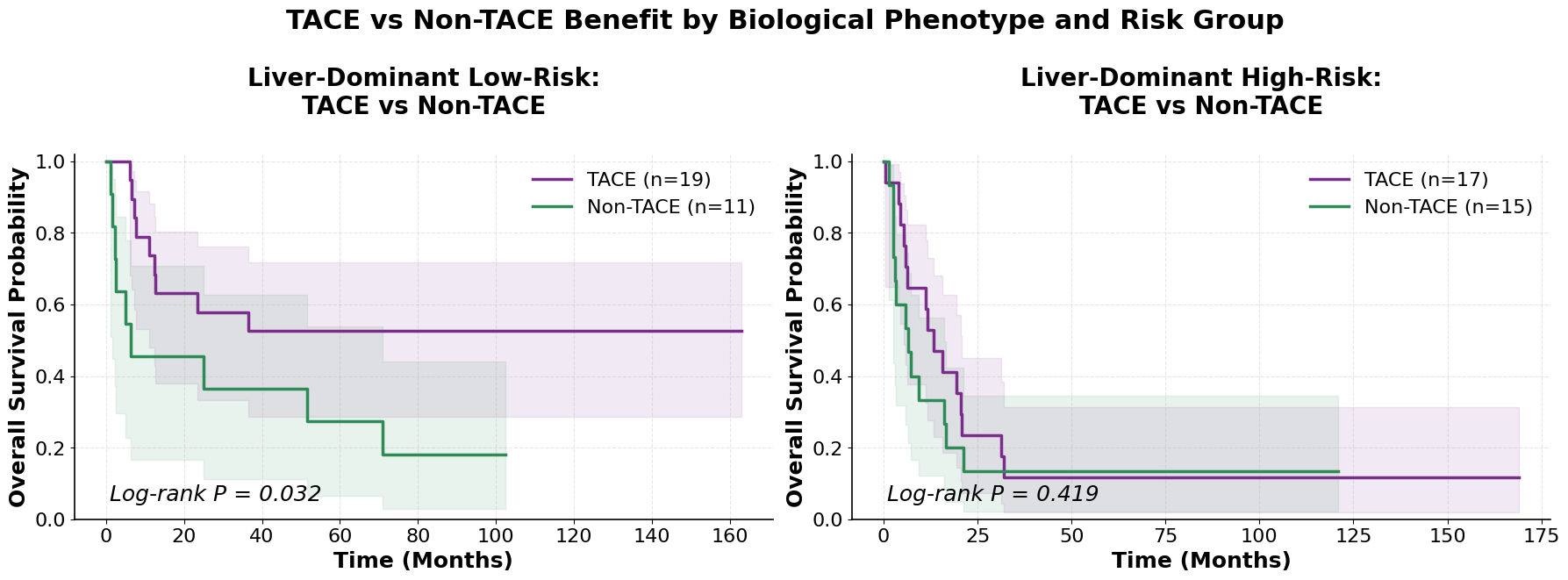}
    
    \caption{In an exploratory treatment analysis, patients identified as liver low-risk by the hepatic pathway showed the greatest predicted benefit from TACE, consistent with clinical expectations.}
\label{fig:treatment_combined}

\end{figure}

BioFact-MoE outperforms all baselines at 12, 18, and 24 months (Table~\ref{tab:comparison}). Based on predicted risk scalar, we can train a decision tree model on the training set to achieve high-risk and low-risk stratification on the test set. In addition, gate weights provide patient-level biological interpretation. Patients are classified as liver-driven ($w_{liver}$ $\geq 0.5$) or tumor-driven based on the dominant expert pathway.
Despite comparable clinical stage, TACE-treated patients in the test set demonstrate marked heterogeneity in survival outcomes. As shown in Fig. \ref{fig:phenotype-aware}, while all patients receiving TACE benefit with log-rank $p{=}$0.024, the liver-dominant phenotype subgroup show improved survival under TACE with log-rank $p{=}$0.009, while the tumor-dominant phenotype subgroup show no difference in terms of survival with log-rank $p{=}$0.834. 

Exploratory treatment analysis is shown in Fig. \ref{fig:treatment_combined}. Hepatic low-risk patients show stronger survival association with TACE, consistent with clinical expectations. In tumor-dominant patients, no significant treatment interaction was detected (all $p > 0.3$); note subgroup sizes in the tumor-dominant TACE and non-TACE arms were limited (all $n \leq 12$ per arm).

\begin{table}[t]
\centering
\caption{Biological validation of pathway specialization by pathway-specific linear probe of available clinical biomarkers.}
\label{tab:bioval}
\setlength{\tabcolsep}{6pt}
\scalebox{0.85}{
\begin{tabular}{lccccccc}
\hline
\textbf{Biomarker} & \textbf{n} & \textbf{Classes} & \textbf{Liver AUC} & \textbf{Tumor AUC} & $\bm{\Delta}$ & $\bm{p}$ & \textbf{Direction}\\
\hline\hline
PALBI & 118 & 3 & 65.4 & 56.4 & 9.00 & <0.05 & Liver > Tumor\\
Immunoscore & 62 & 4 & 54.4 & 59.3  & 4.9 & <0.05 & Tumor > Liver \\
Bilobar disease & 386 & 2 & 59.7 & 63.7 & 4.1 & <0.05 & Tumor > Liver\\
\hline
\end{tabular}}
\end{table}

\subsection{Biological Validation of Pathway Specialization}
\label{sec:bioval}

Table~\ref{tab:bioval} provides evidence that the learned pathway embeddings 
capture biologically meaningful signals aligned with clinical biomarkers. Linear probes (5-fold cross-validation) indicate PALBI, a 
marker of hepatic functional reserve, is more accurately predicted from frozen liver-pathway 
embeddings than tumor-pathway embeddings ($\Delta = 9.0$, $p < 0.05$), confirming 
that the liver expert has specialized toward hepatic biology. Conversely, both tumor-related variables, immunoscore 
and bilobar disease extent, are better captured by 
the tumor-pathway embeddings ($\Delta = 4.9$ and $4.1$ respectively, $p < 0.05$), 
validating that the tumor expert has acquired complementary tumor-specific 
representations. This cross-validated biological alignment provides evidence that BioFact-MoE achieves genuine factorization rather than redundant parallel encoding.

\subsection{Ablation Studies}


Table~\ref{tab:ablation_seg} evaluates the role of biological grounding from both text and images across training and inference.
\textbf{Random Text Split} and \textbf{Swapped Conditioning} disrupt biologically meaningful report assignment, while \textbf{No Text Segmentation} removes textual pathway supervision entirely. \textbf{No Anatomical Masking} eliminates organ-aware visual grounding.  \textbf{Full Report at Inference} keeps pathway-specific text segmentation during 
downstream training but substitutes the full unsegmented report at inference. \noindent\textbf{Full Report} uses the unsegmented report for both 
downstream training and inference. The inductive bias introduced by biological grounding from both text and images is vital for expert specialization in BioFact-MoE. 

\begin{table}[t]
\centering
\caption{Ablation study of biological grounding across training and inference.}
\label{tab:ablation_seg}
\setlength{\tabcolsep}{6pt}
\renewcommand{\arraystretch}{1.2}
\scalebox{0.82}{
\begin{tabular}{llccc}
\hline
\textbf{Stage} & \textbf{Model Variant} 
& \textbf{12-mo AUC} & \textbf{18-mo AUC} & \textbf{24-mo AUC} \\
\hline\hline

Baseline

Pretraining
  & No Text Segmentation            & 71.52 $\pm$ 1.91 & 70.63 $\pm$ 2.43 & 68.98 $\pm$ 3.01 \\

  & Random Text Split               & 68.16 $\pm$ 2.02 & 69.31 $\pm$ 1.64  & 67.84 $\pm$ 0.87\\

  & Swapped Text Conditioning       & 69.28 $\pm$ 2.85 & 70.32 $\pm$ 0.87 & 69.60 $\pm$ 1.79 \\

  & No Anatomical Masking           & 71.67 $\pm$ 2.23 & 72.42 $\pm$ 0.60 & 69.97 $\pm$ 0.71 \\
\hline

Downstream
  & Full Report at Inference & 72.27 $\pm$ 1.47 & 70.19 $\pm$ 2.02 & 68.99 $\pm$ 1.97 \\
  & Full Report        & 74.93 $\pm$ 1.38  & 73.35 $\pm$ 1.11 & 71.42 $\pm$ 1.89 \\
\hline

Ours
  & \textbf{BioFact-MoE}
  & \textbf{75.33 $\pm$ 2.60}
  & \textbf{75.85 $\pm$ 1.45}
  & \textbf{73.96 $\pm$ 3.04} \\
\hline
\end{tabular}}
\end{table}

Table~\ref{tab:ablation} isolates the contribution of each pathway component. 
\textbf{Base Only} removes both pathway-specific experts and retains only the base survival head, establishing 
the performance floor of general image-report fusion without biological 
factorization. \textbf{Liver Only} and \textbf{Tumor Only} retain a single expert pathway and 
remove the other, quantifying the individual prognostic value of hepatic and 
tumor-specific representations. \textbf{Joint Learning (No MoE)} trains liver, tumor, and base 
pathways jointly with equal contribution and no mixture routing, testing whether 
adaptive expert weighting via MoE is necessary beyond simple multi-pathway fusion.

We observe that Tumor \& Base slightly outperforms Liver \& Base, possibly because the shared pathway already captures substantial liver-related global signals, while tumor-specific features benefit more from dedicated specialization. Notably, Liver \& Tumor (No Base) achieves performance comparable to the full BioFact-MoE at 18- and 24-month horizons, with only modest differences at 12 months. This suggests that biologically specialized pathways capture most of the prognostic signal, while the base pathway mainly provides additional robustness rather than driving performance.

Overall, these results indicate that biological factorization is the primary contributor, and the residual MoE design serves to enhance stability rather than act as the main source of performance gain.

\begin{table}[t]
\centering
\caption{Ablation study of pathway modeling contributions.}
\label{tab:ablation}
\setlength{\tabcolsep}{8pt}
\scalebox{0.9}{
\begin{tabular}{lccc}
\hline
\textbf{Model Variant} & \textbf{12-mo AUC} & \textbf{18-mo AUC} & \textbf{24-mo AUC} \\
\hline\hline
Base Only & 65.71 $\pm$ 5.87 & 67.97 $\pm$ 0.64 & 66.91 $\pm$ 2.31 \\
Liver Only &  68.24 $\pm$ 1.00 & 68.85 $\pm$ 1.79 & 67.84 $\pm$ 1.83 \\
Tumor Only & 68.26 $\pm$ 2.03 & 70.28 $\pm$ 0.21 & 69.48 $\pm$ 3.53 \\
Liver \& Base & 68.61 $\pm$ 0.81  & 69.40 $\pm$ 1.62 & 68.51 $\pm$ 2.08 \\
Tumor \& Base & 71.27 $\pm$ 0.12 & 72.70 $\pm$ 0.09 & 70.75 $\pm$ 2.64 \\
Joint Learning (No MoE) & 71.31 $\pm$ 1.64 & 71.65 $\pm$ 0.86 & 71.27 $\pm$ 2.02 \\
Liver \& Tumor (No Base) & 72.71 $\pm$ 0.78 & 75.22 $\pm$ 1.69 & 73.13 $\pm$ 2.22\\
\textbf{BioFact-MoE} & \textbf{75.33 $\pm$ 2.60} & \textbf{75.85 $\pm$ 1.45} & \textbf{73.96 $\pm$ 3.04}  \\
\hline
\end{tabular}}
\end{table}






\section{Conclusion}

We presented BioFact-MoE, a biologically factorized MoE framework 
for vision-language prognostic modeling in hepatocellular carcinoma. By explicitly 
decomposing radiology reports into liver- and tumor-specific pathways and aligning 
pathway-specific LoRA adapters with anatomically masked visual representations, 
BioFact-MoE learns biologically structured multimodal embeddings without requiring 
biomarker supervision. A residual MoE survival head dynamically 
routes patient-specific risk contributions across both pathways, improving survival prediction while enabling phenotype-aware stratification. Biological validation confirms that pathway embeddings align with 
clinically recognized hepatic and tumor markers, supporting our factorization 
as an emergent property of the pretraining design.

Several limitations limited further investigation. First, the LLM-guided report 
decomposition relies on the capability of the LLM; errors in 
pathway assignment may introduce noise into the biological factorization signal. 
Second, structured clinical biomarkers used for biological validation are sparsely 
available, limiting the statistical power of pathway specialization analysis.

%
%
%
\bibliographystyle{splncs04}
\bibliography{mybibliography}
%




\end{document}